# Meteorological time series forecasting based on MLP modelling using heterogeneous transfer functions

C. Voyant[1], ML. Nivet[1], C. Paoli[2], M. Muselli[1] and G. Notton[1]

1- University of Corsica - SPE CNRS UMR 6134, 20250 Corte - France (Corsica)

2- University of Galatasaray - Computer Engineering Department, No:36 34357 Ortaköy, Istanbul - Turkey

voyant@univ-corse.fr

**Abstract**. In this paper, we propose to study four meteorological and seasonal time series coupled with a multi-layer perceptron (MLP) modeling. We chose to combine two transfer functions for the nodes of the hidden layer, and to use a temporal indicator (time index as input) in order to take into account the seasonal aspect of the studied time series. The results of the prediction concern two years of measurements and the learning step, eight independent years. We show that this methodology can improve the accuracy of meteorological data estimation compared to a classical MLP modelling with a homogenous transfer function.

## 1. Introduction

The time series (TS) formalism is useful in many scientific fields [1,3,4]. In the particular case of the meteorology, the prediction is essential to anticipate weather variation and thus to prevent the population of potential risks, but it can also be used for the energy management (intermittent energy sources switching e.g. global solar radiation or wind speed) [5-8]. The measured series are often seasonal and noised; they have repetitive and more or less predictable fluctuations (24 hours and 1 year). The noise increases when the time step between measurements decreases. It is essential to take into account periodicities and to build a predictor with seasonal adjustments [9]. In this paper, we suggest to study periodic meteorological TS measured during 11 years in Ajaccio (France; at an hourly step) and concerning four kinds of data with different degrees of periodicity: solar irradiation (Wh/m$^2$, [6]), humidity (%), ground temperature (◦C) and wind velocity at 10 meters from the ground (Beaufort scale) [10]. The variation coefficients (VC; standard_deviation/mean) of these series are respectively: 1.5, 0.2, 0.4 and 0.5. This coefficient represents the variability and the seasonality of the series; the high values are often related to bad prediction results. In the next section, we will present the one hour ahead prediction methodology based on the heterogeneous transfer functions multilayer perceptron (MLP) and on a particular time index taken as input. Then the results of predictions and the comparison between different architectures will be exposed; finally in the last part, we will analyze the results and will draw conclusions.

## 2. Methods

In this section, we first introduce the global methodology based on the heterogeneous transfer functions MLP that is to say the fact to mix transfer functions within on network [2]. Next, we present the different meteorological data estimation methodologies used during the cross-comparison. .

2.1. MLP with heterogeneous transfer functions

A classical MLP with one hidden layer is, in the time series ($x(t)$) prediction context, defined by equation 1 where $W^1, b^1$ and $W^2, b^2$ are the weights and biases matrices of the hidden and output layers and $f$ the transfer function of the hidden layer (note that in our case, the transfer function of the output layer is the identity function). The tapped delay line ($p$ delays) is defined from the lag operator ($L$) by

$TDL = \{L^{i-1}x(t)\} i \in [1, p]$ considering that $L^i x(t) = x(t - i)$. With this notation, the output of the MLP becomes $O = L^{-1}x(t)$ [3].

$$O = W^2 \, f(W^1 TDL + b^1) + b^2 \qquad (1)$$

Often a hyperbolic tangent function (or similar function) is used for *f*. In this study we propose to combine two types of transfer function in the hidden layer: bijective and non-bijective. As the first one allows activating a specific hidden node once the argument is upper than the bias; the second one allows to introduce an excitation/inhibition balance in the system: low and high signals will have an inhibition effect. We choose to test respectively the hyperbolic tangent function (noted tanh) and its derivative. Figure 1 represents these two functions types and the scaled form of the non-bijective transfer function (expanded between -1 and 1).

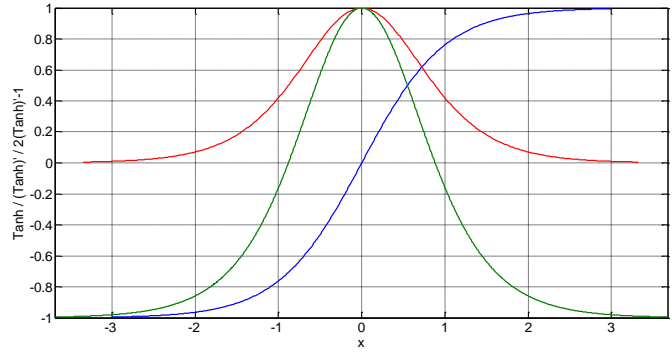

Fig. 1: In blue the tanh function and in red its derivative (= 1-tanh²). The green curve represents the scaled form of (tanh)'

With this modification, the new network is defined by (2), where $W^{1'}$ and $b^{1'}$ define the weights and biases related to the non-bijective nodes.

$$O = W^2(f(W^1 TDL + b^1) + f'(W^{1'} TDL + b^{1'})) + b^2 \qquad (2)$$
$$O = W^2(\tanh(W^1 TDL + b^1) + 1 - \tanh^2(W^{1'} TDL + b^{1'})) + b^2$$

The difference between the "classic" and the heterogeneous transfer functions MLP is related to the two following expressions:

$$\tanh(W^1 TDL + b^1) \text{ and } \tanh(W^1 TDL + b^1) - \tanh^2(W^{1'} TDL + b^{1'}) + 1 \qquad (3)$$

If we pose $W^{1'} TDL + b^{1'} = (W^1 TDL + b^1)\Delta$ where $\Delta$ is the asymmetry rate, and $W^1 TDL + b^1 = u$, we obtain that in the classic case the output is related to $\tanh(u)$ while in the heterogeneous transfer functions MLP case it is related to $\tanh(u) - \tanh^2(u\Delta) + 1$. In figure 2 are shown these two types of outputs related to linear combination of inputs (*u*) and to $\Delta$. If $\Delta$ is upper than 0.5 then the global output is non-bijective and the inhibition phenomenon becomes leading. Concerning the low values of $\Delta$ the output is not between -1 and 1 but between -0.3 and 1.7 (shift of 0.7).

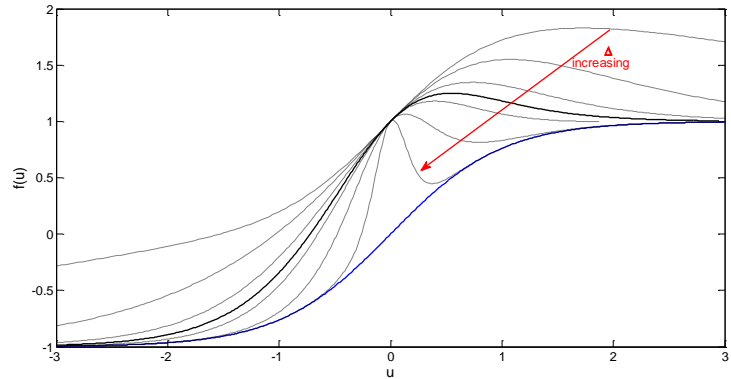

Fig. 2: Output of the classic (blue) and the heterogeneous transfer functions MLP (dotted curve) relating to the *u* parameter (linear combination of inputs) and the asymmetry rate Δ (0.2, 0.5, 0.8, 1.2, 2 and 5). In black, the output concerning Δ=1.

2.2. Meteorological data forecast

The test and validation process of the proposed methodology consists in a cross-comparison with some classical tools usually proposed in literature in order to make Meteorological data forecasting. These different methodologies of prediction are tested through the two last years of measurements, all the others data are related to learning step of the MLP.

*2.2.1. Persistence*

The first studied prediction methods is the persistence method; the simplest way of producing a forecast. The persistence method assumes that the actual conditions will not change. To improve the results and to take into account the fact that the TS is more or less periodical, we corrected the persistence with a scale term computed from the moving average related to the nine first years [4,9].

*2.2.2. MLP*

In the presented study, the MLPs have been computed with the Matlab© software and its Neural Network toolbox. The characteristics chosen related to previous works are the following: one hidden layer, the activation functions are the continuously and differentiable hyperbolic tangent (hidden) and linear (output), the Levenberg-Marquardt (approximation to the Newton's method) learning algorithm with a maximum fail parameter before stopping training equal to 3. The optimization of the number of input nodes is done with the partial autocorrelation factor, which measures the degree of association between two measures, with the effect of a set of controlling intermediate measures removed. Concerning the four types of TS considered, this number varies between 4 and 9 lags. The number of hidden neurons is taken equal to the input nodes number. The results shown in the following parts are related to the best network among six different trainings coupled with a random weight initialization [1,4,7]. Concerning the MLP with heterogeneous transfer functions, the characteristics are described in the section 2.1.

*2.2.3. MLP with a time index input*

The last approach consists in adding a time index input to the two MLP previously proposed, that is to say, the classical one and the MLP with the heterogeneous transfer function. An input is added and is related to the time concerning the hour of prediction. This node takes the values i/24 (for i=1 to 24) for each hour of the day [9].

**3. Results**

The first series of results (Table 1) is related to the persistence (P), normal MLP (noted N-MLP) and the heterogeneous transfer functions MLP (noted HTF-MLP; 50% of tanh' transfer function in the hidden layer). The error metric used is the nRMSE (RMSE/mean). Concerning the use of the MLP, the results presented in the following tables are related to the runs (among six) minimizing the nRMSE.

|  | P | N-MLP | HTF-MLP |
|---|---|---|---|
| Humidity | **0.070** | 0.073 | 0.073 |
| Solar irradiation | 0.319 | **0.305** | 0.308 |
| Temperature | **0.058** | 0.067 | 0.067 |
| Wind speed | 0.394 | **0.359** | **0.359** |

Table. 1: nRMSE related to the persistence, normal MLP and heterogeneous transfer functions MLP. In bold the lowest nRMSE.

In Table 1, we see that HTF-MLP is not really efficient, it gives the best results only for only one TS (Wind speed) tied with N-MLP. With a MLP, one way to improve the results is often to make the time series stationary and to operate a preprocessing before to choose the inputs nodes. In our case, there are four kinds of data, so we chose to apply a seasonal adjustment by periodic coefficients (computed from moving average during the learning phase). In the following results, when this pre-post processing is used, the suffix –s is added (example N-MLP-s). The Table 2 shows the impact of this processing on the nRMSE value.

|  | N-MLP-s | HTF-MLP-s |
|---|---|---|
| Humidity | **0.070** | **0.070** |
| Solar irradiation | **0.333** | 0.334 |
| Temperature | **0.059** | **0.059** |
| Wind speed | 0.351 | **0.350** |

Table. 2: nRMSE related to the normal MLP and heterogeneous transfer functions MLP and concerning the time series made stationary. In bold the lowest nRMSE.

The stationary process doesn't improve systematically the nRMSE, in fact, in the solar radiation case the error is increased by more than 2 percentage point. The second tool that should improve results is the time index taken as input of the networks. In the following parts, we will note –t (example N-MLP-s-t) when this index is used. The Table 3 shows the impact of this exogenous data (multivariate analysis).

|  | N-MLP-t | HTF-MLP-t | N-MLP-s-t | HTF-MLP-s-t |
|---|---|---|---|---|
| Humidity | **0.069** | 0.070 | 0.071 | 0.070 |
| Solar radiation | **0.292** | 0.298 | 0.324 | 0.331 |
| Temperature | 0.058 | 0.060 | 0.059 | **0.057** |
| Wind speed | 0.358 | 0.357 | **0.351** | 0.353 |

Table. 3: nRMSE related to the normal MLP and heterogeneous transfer functions MLP and concerning the time index taken as input. In bold the lowest nRMSE.

Considering the different configurations -s, -t, non-bijective or bijective transfer functions of hidden nodes, a lot of combinations are possible. The next table (4) summarizes for each data the predictor with the lowest nRMSE and the related error metric.

*the hidden node with the (tanh)' transfer function is only linked to the time index input

|  | VC | Type | nRMSE | Time index | Stationary | Transfer functions |
|---|---|---|---|---|---|---|
| Humidity | 0.2 | N-MLP-t | 0.069 | Yes | No | 100% tanh |
| Solar irradiation | 1.5 | N-MLP-t | 0.292 | Yes | No | 100% (tanh)' |
| Temperature | 0.4 | HTF-MLP-s-t | 0.057 | Yes | Yes | 50% tanh 50% (tanh)' |
| Wind speed | 0.5 | HTF-MLP-s | 0.350 | No | Yes | 80% tanh 20% (tanh)'* |

Table. 3: Best configuration considering nRMSE, for all combinations of tested MLP parameters.

The time index and the (tanh)' transfer function are the tools giving the best results 75% of the cases). The stationary process does not define really the best approaches. For the two time series with the lowest VC the nRMSE is under 10% whereas for the other series (solar irradiation and wind speed) the nRMSE is close to 30%.

## 4. Conclusion

In this study, a MLP coupling two transfer functions, a time index and a stationary process was studied. If the interest of the time index taken as inputs of the network is proved in this study, future studies expanding the nature or the location related to time series will be necessary to show the interest of the MLP with heterogeneous transfer functions. As conclusion of this paper, for each case concerning the time series forecasting of meteorological data, the possibility to use a time index, a stationary process and an inhibition transfer functions should be considered.

**References**

[1] M. A. Behrang, E. Assareh, A. Ghanbarzadeh and A. R. Noghrehabadi, The potential of different artificial neural network (ANN) techniques in daily global solar radiation modeling based on meteorological data, Solar Energy, 84:1468–1480, Elsevier, 2010.

[2] W. Duch and N. Jankowski, Transfer functions: hidden possibilities for better neural networks, ESANN'2001 proceedings - European Symposium on Artificial Neural Networks, Bruges (Belgium), 25-27 April 2001, D-Facto public., ISBN 2-930307-01-3, pp. 81-94

[3] S.F. Crone, Stepwise selection of artificial neural networks models for time series prediction, J. Intelligent Syst. 14: 99–122, De Gruyter, 2005.

[4] J. G. De Gooijer and R. J. Hyndman, 25 years of time series forecasting, Int. J. Forecasting, 22: 443–473, Elsevier, 2006.

[5] C. Duchon and R. Hale, Time Series Analysis in Meteorology and Climatology: An Introduction, Wiley-Blackwell, 2012.

[6] F. O. Hocaoglu, Stochastic approach for daily solar radiation modeling, Solar Energy, 85: 278–287, Elsevier, 2011.

[7] S. Kalogirou, Artificial neural networks in renewable energy systems applications: A review, Renew. Sustain. Energy Rev., 5: 373–401, Elsevier, 2001.

[8] A. Mellit and S. Kalogirou, Artificial intelligence techniques for photovoltaic applications: A review, Prog. Energy Combustion Sci., 34: 574–632, Elsevier, 2008.

[9] C. Voyant, C. Paoli, M. Muselli and M.-L. Nivet, Multi-horizon solar radiation forecasting for Mediterranean locations using time series models, Renew. Sustain. Energy Rev., 28: 44–52, Elsevier, 2013.

[10] T.T. Warner, Numerical Weather and Climate Prediction, Cambridge University Press, 2010.